\documentclass[11pt,a4paper]{article}

\usepackage[hyperref]{naaclhlt2019}

\usepackage{times}
\usepackage{latexsym}
\usepackage{times}
\usepackage{latexsym}
\usepackage{url}
\usepackage{algorithm2e}
\usepackage{graphicx}
\usepackage{float}
\usepackage{array}
\usepackage{pgfplots}
\usepackage{tikz}
\usepackage{tikz-qtree}
\usepackage{amssymb}
\usepackage{amsthm}
\usepackage{amsmath}
\usepackage{booktabs}
\usepackage{xcolor}

\aclfinalcopy 
\def\aclpaperid{} 

\newcommand{\word}[1]{\emph{#1}}
\newcommand{\tech}[1]{\emph{#1}}

\DeclareMathOperator{\relu}{relu}

\definecolor{roamdarkblue}{HTML}{0499CC}
\definecolor{roamlightblue}{HTML}{03A9F4}
\definecolor{roamdarkgray}{HTML}{838A8A}
\definecolor{roamlightgray}{HTML}{B8B8B8}
\definecolor{roamgreen}{HTML}{4D8951}
\definecolor{roamblack}{HTML}{212121}
\definecolor{roamsteelblue}{HTML}{9BB8D7}
\definecolor{roamorange}{HTML}{FDBA58}
\definecolor{roamwhite}{HTML}{FAFAFA}
\definecolor{roampurple}{HTML}{876DB5}
\definecolor{mymaroon}{HTML}{881C1c}

\title{Stress-Testing Neural Models of Natural Language Inference with
  Multiply-Quantified Sentences}

\author{Atticus Geiger \\
  Stanford Symbolic Systems Program \\
  {\tt atticusg@stanford.edu} \\[1ex]
    \textbf{Lauri Karttunen} \\
  Stanford Linguistics \\
  {\tt laurik@stanford.edu}\\\And
  Ignacio Cases \\
  Stanford Linguistics \\
  {\tt cases@stanford.edu}\\[1ex]
  \textbf{Christopher Potts} \\
  Stanford Linguistics \\
  {\tt cgpotts@stanford.edu} \\}

\date{}

\begin{document}
\maketitle

\begin{abstract}
  Standard evaluations of deep learning models for semantics
  using naturalistic corpora are limited in what they can tell us
  about the fidelity of the learned representations, because the
  corpora rarely come with good measures of semantic
  complexity. To overcome this limitation, we present a method
  for generating data sets of multiply-quantified natural language
  inference (NLI) examples in which semantic complexity can be
  precisely characterized, and we use this method to show that a
  variety of common architectures for NLI inevitably fail to encode
  crucial information; only a model with forced lexical alignments
  avoids this damaging information loss.
\end{abstract}

\section{Introduction}\label{sec:introduction}


Deep learning approaches to semantics hypothesize that it is feasible
to learn fixed-dimensional distributed representations that encode the
meanings of arbitrarily complex sentences. Well-known mathematical
results show that this hypothesis is correct in terms of
representational capacity \citep{Cybenko:1989}, but it remains an
empirical question whether a given model architecture can achieve the
desired representations in practice. In addressing this question,
researchers generally rely on corpora of naturalistic examples, using
comparative performance metrics as a proxy for the underlying capacity
of the models to learn rich meaning representations. However, these
corpora rarely come with good independent measures for semantic
complexity (but see \citealt{Williams-etal:2017}), so they too leave
us guessing about precisely what has been learned by any given system.

This paper presents a method for generating artificial data sets in
which the semantic complexity of individual examples can be precisely
characterized. Our task is natural language inference (NLI). There are
diverse, high-quality naturalistic corpora for this task,
and a variety of architectures have been evaluated on them, with some
clear success. However, the corpora themselves have been scrutinized
and found to contain patterns that lower our confidence that the models
are learning robust semantic representations
\citep{Rudinger-etal:2017,Poliak-etal:2018, Glockner:2018,Gururangan:2018,Tsuchiya:2018}.

We propose to pair these evaluations with ones conducted on our data
sets to achieve a fuller picture. Our method is built around an
interpreted formal grammar that generates sentences containing
multiple quantifiers, modifiers, and negations. We constrain the
open-domain vocabulary to ensure that all items neither entail nor
exclude all others, to trivialize their contributions. This stresses
models with learning interactions between logically complex function
words. The sentences from this grammar are
deterministically translated into first-order logic, and an off-the-shelf
theorem prover is used to generate NLI examples. In this setting, we have
control over all aspects of the generated data set and complete visibility
into how different systems handle specific classes of example.

\begin{figure*}
  \centering
  \input{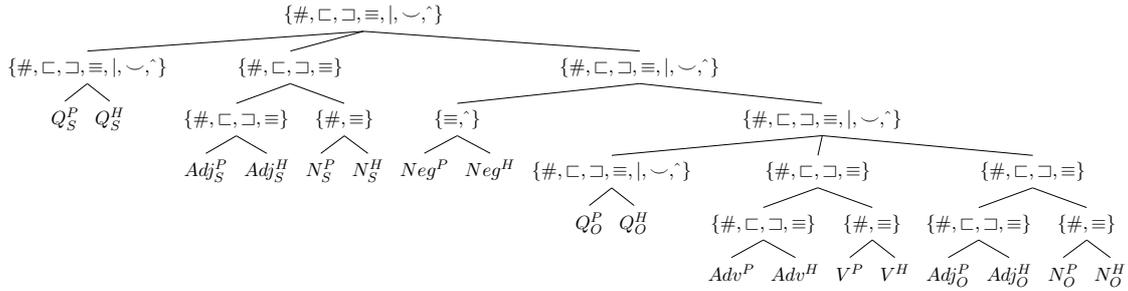}
  \caption{An aligned NLI example. At each non-terminal node,
    we show the full range of semantic relations that can hold
    between the two aligned phrases given the constraints we
    impose on these examples. We use the seven basic semantic
    relations of \newcite{MacCartney:09}: $\#$ = independence; $\sqsubset$ =
    entailment; $\sqsupset$ = reverse entailment; $|$ = alternation;
    $\smile$ = cover; $\hat{} \:$ = negation; $\equiv$ =
    equivalence. }
  \label{fig:bigtree}
\end{figure*}

We evaluate a number of different architectures,
including LSTM sequence models with attention, tree-structured neural
networks (TreeNN), and a tree-structured neural network that processes
aligned NLI examples (CompTreeNN). Our central finding is that only
the CompTreeNN performs perfectly. Even the LSTM with attention, which
has the space to learn lexical alignments, fails to find optimal
solutions. What is special about the CompTreeNN is its ability to abstract
away the identity of specific lexical items by computing and propagating
lexical semantic relations. When stressed with
even our small fragment of natural language's complexity, the other models lose the identity
of open-class lexical items as the sentence embeddings are recursively
constructed. This results in systematic errors, calling into
question the viability of these models for semantics.

\section{Data Generation}\label{sec:data}

Our fragment $G$ consists of sentences of the form

\newcommand{\nlvar}[2]{\ensuremath{\text{#1}_{\text{#2}}}}
\begin{center}
  \nlvar{Q}{S} \nlvar{Adj}{S} \nlvar{N}{S} \nlvar{Neg}{} \nlvar{Adv}{}
  \nlvar{V}{} \nlvar{Q}{O} \nlvar{Adj}{O} \nlvar{Adj}{O} \nlvar{N}{O}
\end{center}
where \nlvar{N}{S} and \nlvar{N}{O} are nouns, \nlvar{V}{} is a verb,
\nlvar{Adj}{S} and \nlvar{Adj}{O} are adjectives, and \nlvar{Adv}{} is
an adverb. \nlvar{Neg}{} is \word{does not}, and \nlvar{Q}{S} and
\nlvar{Q}{O} can be \word{every}, \word{not every}, \word{some}, or
\word{no}; in each of the remaining categories, there are 100 words.
Additionally, \nlvar{Adj}{S}, \nlvar{Adj}{O}, \nlvar{Adv}{}, and
\nlvar{Neg}{} can be the empty string, which is represented in the data
by a unique token. Semantic scope is fixed by surface order,
with earlier elements scoping over later ones.

For NLI, we define the set of premise--hypothesis pairs
$\mathcal{S} \subset G \times G$ such that
$(s_{p}, s_{h}) \in \mathcal{S}$ iff the non-identical non-empty
nouns, adjectives, verbs, and adverbs with identical positions in
$s_{p}$ and $s_{h}$ are mutually consistent (semantic independence).
This constraint on $\mathcal{S}$ trivializes the task of determining
the lexical relations between adjectives, nouns, adverbs, and verbs, since
the relation is equality where the two aligned elements are identical,
otherwise independence. Furthermore, it follows that distinguishing
contradictions from entailments is trivial. The only sources of
contradictions are negation and the negative quantifiers \word{no} and
\word{not every}. Consider $(s_{p}, s_{h}) \in \mathcal{S}$ and let
$C$ be the number of times negation or a negative quantifier occurs
$s_{p}$ and $s_{h}$. If $s_{p}$ contradicts $s_{h}$, then $C$ is odd;
if $s_{p}$ entails $s_h$, then $C$ is even.

Figure~\ref{fig:bigtree} summarizes the range of relations that
can hold between aligned pairs of subconstituents given the
constraints we impose on $\mathcal{S}$. Even with the contributions of
open-class lexical items trivialized, the level of complexity remains
high, and all of it emerges from semantic composition, rather than
from lexical relations.

Our corpora use the  three-way labeling scheme of
\tech{entailment}, \tech{contradiction}, and \tech{neutral}. To assign
these labels, we translate each premise--hypothesis pair into
first-order logic and use Prover9 \cite{McCune:2005}. We assume no
expression is empty or universal and encode these assumptions as
additional premises. This label generation
process implicitly assumes the relation between unequal artificial
words is independence.

For NLI corpora, we create samples from $\mathcal{S}$ in which, for a
given example, every adjective--noun and adverb--verb pair across the
premise and hypothesis is equally likely to have the relation
equality, subset, superset, or independence. Without this balancing,
any given adjective--noun and adverb--verb pair across the premise and
hypothesis has more than a 99\% chance of being in the
independence relation. Even with this step, 98\% of the pairs are neutral,
so we again sample to create corpora that are
balanced across the NLI labels.\footnote{Our data set generation code: \url{https://github.com/atticusg/MultiplyQuantifiedData}}

Our approach to data set generation mostly closely resembles that of
\citet{Bowman2:15}, who conduct a range of artificial language
experiments. In one, they create corpora of single-quantified examples
and show that simple tree-structured neural networks are able to learn
the semantic relations between them. While \citeauthor{Bowman2:15} use
a larger set of quantifiers than we do, their syntactic frames are
much simpler, their lexicon has only nine open-class items, and the
lexical items have complex semantic relationships to each other.
Our data sets are thus much more complex, which is reflected in
the results discussed later.

\begin{table*}
  \small
  \centering
  \setlength{\tabcolsep}{14pt}
\begin{tabular}{r c c c c}
  \toprule
  Model & Train & Dev & Test & Informative Open-class Subset  \\
  \midrule
  CBoW           & $69.18 \pm 0.75$ & $69.91 \pm 0.27$ & $69.66 \pm 0.23$ & $82.30 \pm 1.20$\\
  LSTM Encoder   &$96.05 \pm 0.29$ & $95.83 \pm 0.14$ & $95.61 \pm 0.21$ & $26.90 \pm 1.44$\\
  TreeNN         & $96.20 \pm 0.17$ & $96.19 \pm 0.15$ & $95.99 \pm 0.11$ & $31.09 \pm 2.91$\\
  Attention LSTM &$97.50 \pm 2.69$ & $95.98 \pm 2.23$ & $95.82 \pm 2.16$ & $35.69 \pm 35.98$\\
  CompTreeNN     &$99.85 \pm 0.07$ & $99.87 \pm 0.06$ & $99.85 \pm 0.12$ & $98.05 \pm 1.02$ \\
  \bottomrule
\end{tabular}


  \caption{Mean accuracy of 5 runs, with 95\% confidence intervals. `Informative Open-class Subset' are the Test set examples labeled neutral solely due to the independence relation between open-class
  lexical items.}
  \label{tab:results}
\end{table*}

\section{Models}\label{sec:models}

We consider five different model architectures:
\begin{description}

\item[CBoW] Premise and hypothesis are represented by the average of
  their respective word embeddings (continuous bag of words).

\item[LSTM] Premise and hypothesis are processed as
  sequences of words using a recurrent neural network (RNN;
  \citealt{Elman:1990}) with LSTM
  cells \citep{Hochreiter:97}, and the final hidden state of each
  serves as its representation \citep{Bowman:15}.

\item[TreeNN] Premise and hypothesis are processed as trees,
  and the semantic compsition function is a single layer feed
  forward network \cite{Socher:2011,Socher2:2011}. The value of
  the root node is the semantic representation in each case.

\item[Attention LSTM] An LSTM RNN
  with word-by-word attention \citep{Rock:15}.

\item[CompTreeNN] Premise and hypothesis are processed as a
  single aligned tree, as in figure~\ref{fig:bigtree}. This
  model is inspired by research in natural logic
  \citep{MacCartney09,Icard:Moss:2013:LILT,Bowman2:15}.

\end{description}

For the first three models, the premise and hypothesis representations
are concatenated. For the CompTreeNN and  LSTM with attention,
there is just a single representation of the pair. In all cases, the
premise-hypothesis representation is fed through two hidden layers and
a softmax layer.

All models are initialized with random 100-dimensional word vectors and
optimized using Adam \cite{pennington:14,Kingma:14}. A grid
hyperparameter search was run over dropout values of $[0,0.1,0.2]$
on the output and keep layers of LSTM cells, learning rates of
$[1\mathrm{e}{-2},3\mathrm{e}{-4},1\mathrm{e}{-3}]$, L2 regularization
values of $[0,1\mathrm{e}{-4},1\mathrm{e}{-3}]$ on all weights, and
activation functions $\relu$ and $\tanh$.

\begin{figure}
  \centering
  \begin{tikzpicture}[scale=0.7, line width=2mm]
\begin{axis}[
    title={},
    no markers,
    every axis plot/.append style={ultra thick},
    xlabel={\# of training examples},
    ylabel={Accuracy},
    xmin=0, xmax=5000,
    ymin=35, ymax=100,
    xtick={0,1000,2000,3000,4000,5000,6000},
    xticklabels={0,1M,2M,3M,4M,5M,6M},
    ytick={40,50,60,70,80,90},
    legend style={at={(0.95,0.48)}}
]

\addplot[
    color=roamdarkblue,
       error bars/.cd,
    y dir=both,
    y explicit
    ]
    coordinates {
(0,33.3333333333333)(100, 91.98919891989199) (200, 94.1994199419942) (300, 94.2994299429943) (400, 94.28942894289429) (500, 94.34943494349434) (600, 94.41944194419442) (700, 94.40944094409441) (800, 94.47944794479449) (900, 94.52945294529454) (1000, 94.72947294729474) (1100, 95.02950295029503) (1200, 95.14951495149515) (1300, 95.10951095109512) (1400, 95.33953395339535) (1500, 95.26952695269527) (1600, 95.3095309530953) (1700, 95.3095309530953) (1800, 95.4095409540954) (1900, 95.31953195319532) (2000, 95.47954795479548) (2100, 95.66956695669568) (2200, 95.53955395539553) (2300, 95.63956395639563) (2400, 95.54955495549555) (2500, 95.73957395739573) (2600, 95.7095709570957) (2700, 95.53955395539553) (2800, 95.57955795579558) (2900, 95.63956395639563) (3000, 95.64956495649565) (3100, 95.68956895689568) (3200, 95.73957395739573) (3300, 95.5895589558956) (3400, 95.73957395739573) (3500, 95.42954295429543) (3600, 95.73957395739573) (3700, 95.65956595659566) (3800, 95.68956895689568) (3900, 95.71957195719571) (4000, 95.65956595659566) (4100, 95.62956295629563) (4200, 95.61956195619562) (4300, 95.72957295729573) (4400, 95.73957395739573) (4500, 95.74957495749575) (4600, 95.68956895689568) (4700, 95.77957795779578) (4800, 95.74957495749575) (4900, 95.79957995799579) (5000, 95.72957295729573) (5100, 95.71957195719571) (5200, 95.80958095809581) (5300, 95.83958395839583) (5400, 95.71957195719571) (5500, 95.87958795879588) (5600, 95.77957795779578) (5700, 95.74957495749575) (5800, 95.75957595759576) (5900, 95.81958195819583)
    };
    \addlegendentry{Tree RNN}

\addplot[
    color=roamgreen,
       error bars/.cd,
    y dir=both,
    y explicit
    ]
    coordinates {
(0,33.3333333333333)(100, 66.06660666066607) (200, 72.47724772477248) (300, 74.85748574857486) (400, 77.57775777577758) (500, 78.37783778377838) (600, 78.42784278427843) (700, 78.33783378337834) (800, 79.05790579057906) (900, 78.47784778477848) (1000, 78.78787878787878) (1100, 78.7078707870787) (1200, 78.8878887888789) (1300, 78.16781678167817) (1400, 79.25792579257926) (1500, 78.95789578957896) (1600, 78.86788678867886) (1700, 78.86788678867886) (1800, 78.5978597859786) (1900, 78.75787578757875) (2000, 78.83788378837883) (2100, 78.4878487848785) (2200, 79.007900790079) (2300, 78.84788478847885) (2400, 78.83788378837883) (2500, 78.8078807880788) (2600, 78.8878887888789) (2700, 79.05790579057906) (2800, 79.06790679067906) (2900, 78.81788178817882) (3000, 78.83788378837883) (3100, 79.11791179117913) (3200, 79.26792679267926) (3300, 84.34843484348434) (3400, 95.84958495849585) (3500, 98.48984898489849) (3600, 99.22992299229924) (3700, 98.60986098609861) (3800, 98.90989098909891) (3900, 99.32993299329934) (4000, 99.15991599159916) (4100, 99.2999299929993) (4200, 99.34993499349935) (4300, 99.33993399339934) (4400, 99.44994499449945) (4500, 99.40994099409941) (4600, 99.38993899389939) (4700, 99.38993899389939) (4800, 99.46994699469947) (4900, 99.47994799479947) (5000, 99.2999299929993) (5100, 99.45994599459947) (5200, 99.51995199519952) (5300, 99.58995899589958) (5400, 99.57995799579959) (5500, 99.55995599559955) (5600, 99.54995499549955) (5700, 99.42994299429942) (5800, 99.57995799579959) (5900, 99.58995899589958)
    };
     \addlegendentry{CompTreeNN}

\addplot[
    color=mymaroon,
       error bars/.cd,
    y dir=both,
    y explicit
    ]
    coordinates {
(0,33.3333333333333)(100, 53.565356535653564) (200, 60.436043604360435) (300, 93.41934193419343) (400, 94.00940094009401) (500, 94.14941494149414) (600, 94.32943294329434) (700, 94.25942594259425) (800, 94.44944494449446) (900, 94.47944794479449) (1000, 94.4994499449945) (1100, 94.44944494449446) (1200, 94.43944394439444) (1300, 94.48944894489449) (1400, 94.50945094509451) (1500, 94.46944694469447) (1600, 94.47944794479449) (1700, 94.3994399439944) (1800, 94.48944894489449) (1900, 94.47944794479449) (2000, 94.4994499449945) (2100, 94.3994399439944) (2200, 94.3994399439944) (2300, 94.66946694669467) (2400, 95.16951695169517) (2500, 95.45954595459546) (2600, 95.28952895289528) (2700, 95.56955695569556) (2800, 95.5895589558956) (2900, 95.61956195619562) (3000, 95.4995499549955) (3100, 95.4995499549955) (3200, 95.5895589558956) (3300, 95.86958695869588) (3400, 95.77957795779578) (3500, 95.81958195819583) (3600, 95.83958395839583) (3700, 95.87958795879588) (3800, 95.82958295829583) (3900, 95.79957995799579) (4000, 95.76957695769576) (4100, 95.73957395739573) (4200, 95.87958795879588) (4300, 95.62956295629563) (4400, 95.81958195819583) (4500, 95.76957695769576) (4600, 95.74957495749575) (4700, 95.68956895689568) (4800, 95.67956795679568) (4900, 95.7095709570957) (5000, 95.74957495749575) (5100, 95.80958095809581) (5200, 95.7095709570957) (5300, 95.68956895689568) (5400, 95.77957795779578) (5500, 95.77957795779578) (5600, 95.74957495749575) (5700, 95.74957495749575) (5800, 95.5095509550955) (5900, 95.67956795679568)
    };
    \addlegendentry{LSTM}

\addplot[
    color=roampurple,
       error bars/.cd,
    y dir=both,
    y explicit
    ]
    coordinates {
(0,33.3333333333333)(100, 48.56485648564857) (200, 47.94479447944794) (300, 53.135313531353134) (400, 53.055305530553056) (500, 52.92529252925292) (600, 54.525452545254524) (700, 54.825482548254826) (800, 56.405640564056405) (900, 58.3058305830583) (1000, 62.946294629462955) (1100, 64.12641264126412) (1200, 66.25662566256626) (1300, 67.96679667966797) (1400, 67.89678967896789) (1500, 67.95679567956796) (1600, 68.43684368436843) (1700, 68.33683368336834) (1800, 68.22682268226824) (1900, 68.31683168316832) (2000, 68.5968596859686) (2100, 68.43684368436843) (2200, 68.46684668466847) (2300, 68.5968596859686) (2400, 68.74687468746875) (2500, 68.94689468946895) (2600, 68.8068806880688) (2700, 68.84688468846885) (2800, 68.78687868786879) (2900, 68.73687368736874) (3000, 68.78687868786879) (3100, 68.6968696869687) (3200, 69.04690469046905) (3300, 68.8968896889689) (3400, 69.08690869086908) (3500, 69.04690469046905) (3600, 69.24692469246925) (3700, 68.95689568956895) (3800, 69.2969296929693) (3900, 69.44694469446945) (4000, 68.65686568656866) (4100, 69.006900690069) (4200, 69.03690369036903) (4300, 69.55695569556956) (4400, 69.05690569056905) (4500, 69.10691069106912) (4600, 69.23692369236923) (4700, 69.27692769276928) (4800, 69.33693369336935) (4900, 69.33693369336935) (5000, 68.83688368836883) (5100, 68.97689768976898) (5200, 69.14691469146915) (5300, 68.70687068706872) (5400, 69.03690369036903)
    };
     \addlegendentry{CBoW}

\addplot[
    color=roamorange,
       error bars/.cd,
    y dir=both,
    y explicit
    ]
    coordinates {
(0,33.3333333333333)(100, 70.997099709971) (200, 93.92939293929393) (300, 94.13941394139414) (400, 93.76937693769378) (500, 94.36943694369437) (600, 94.45944594459445) (700, 93.90939093909391) (800, 94.32943294329434) (900, 94.43944394439444) (1000, 94.35943594359436) (1100, 94.45944594459445) (1200, 94.45944594459445) (1300, 94.36943694369437) (1400, 94.47944794479449) (1500, 93.42934293429343) (1600, 94.43944394439444) (1700, 94.50945094509451) (1800, 94.4994499449945) (1900, 94.50945094509451) (2000, 94.48944894489449) (2100, 94.45944594459445) (2200, 94.48944894489449) (2300, 94.50945094509451) (2400, 94.48944894489449) (2500, 94.42944294429442) (2600, 94.48944894489449) (2700, 94.46944694469447) (2800, 94.48944894489449) (2900, 94.45944594459445) (3000, 94.50945094509451) (3100, 94.50945094509451) (3200, 94.4994499449945) (3300, 94.50945094509451) (3400, 94.45944594459445) (3500, 94.50945094509451) (3600, 93.33933393339335) (3700, 94.27942794279429) (3800, 94.51945194519452) (3900, 94.4994499449945) (4000, 94.48944894489449) (4100, 94.51945194519452) (4200, 94.4994499449945) (4300, 94.4994499449945) (4400, 94.50945094509451) (4500, 94.50945094509451) (4600, 94.50945094509451) (4700, 94.50945094509451) (4800, 94.48944894489449) (4900, 94.50945094509451) (5000, 91.74917491749174) (5100, 94.36943694369437) (5200, 94.50945094509451) (5300, 94.50945094509451) (5400, 94.50945094509451) (5500, 94.4994499449945) (5600, 94.41944194419442) (5700, 94.50945094509451) (5800, 94.4994499449945) (5900, 94.4994499449945)
    };
     \addlegendentry{Attention LSTM}

\end{axis}
\end{tikzpicture}
  \caption{Model performance throughout training.}
  \label{fig:epochs}
\end{figure}
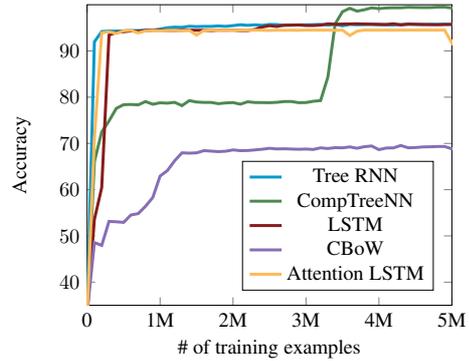

\section{Results and Analysis}\label{sec:results}

\subsection{Overall Results}

Table~\ref{tab:results} summarizes the performance of each model
trained on 500K examples and tested on a disjoint set of 10K examples.
Figure~\ref{fig:epochs} shows dev-set model performance by training
epoch. The CBoW begins far behind all the other models and never
catches up. The LSTM, Attention LSTM, and TreeNN both jump quickly to
${\approx}94\%$, and slightly increase performance plateauing at very
good but not perfect performance (${\approx} 96\%$). When trained further,
the train set is overfit and test performance declines to ${\approx}94\%$.
Only the CompTreeNN is able to perform perfectly.

\subsection{A Shared Suboptimal Solution}

We noted above that the LSTM, Attention LSTM, and TreeNN models get
stuck at ${\approx} 94\%$ accuracy early in training. Because of the highly
controlled way that we generate our data sets, we can pinpoint exactly
why this happens. These models find the same suboptimal solution: they
learn the identity and order of quantifiers, the importance of negation,
and whether or not adjectives and adverbs are empty, but they are unable
to make use of the specific identity of non-empty nouns, verbs, adjectives,
and adverbs. As a result, they make systematic errors on neutral examples
that would be contradictions or entailments if aligned open-class lexical
were equal (see Table \ref{tab:results}). The following collapsing is
performed:

\noindent
  \word{Every Swiss baker madly rubs some rock} \ $\Rightarrow$ \\
  \phantom{-}\hfill \textit{every} $Adj_S$ $N_S$ \textit{Adv V some} $N_O$

\noindent
  \word{Every wild baker sells some rock} \ $\Rightarrow$ \\
  \phantom{-}\hfill \textit{every} $Adj_S$ $N_S$ \textit{ V some} $N_O$

As a result of this collapsing, this looks like an entailment
relation, because the only difference is the deletion of the
adverb, which expands the scope of the universal quantification.

For the Encoder LSTM and TreeNN models, there is a natural explanation
for why these errors are made. These models separately bottleneck the
information from the premise and hypothesis into two 100-dimensional
embeddings before a prediction is made using the concatenation of
these embeddings. The function words are, like function words in
natural data, very complex and very frequent as compared to open-class
items. The stress of learning them seems to nullify the models' ability
to store the precise identity of up to six open-class items per example,
each drawn from a lexicon of 100.
Both these models make minor improvements on this solution to achieve
${\approx} 96 \%$. The LSTM sometimes notices when object
nouns and adjectives differ across the premise and hypothesis, and the
TreeNN model is able to do so with subject nouns and adjectives.
These are precisely the lexical items whose contributions to the sentence
embeddings are most recent, emphasizing the architectural nature of this
problem.

The performance of the Attention LSTM has a high variance. On some runs, it
gets stuck at ${\approx}94\%$ test accuracy, and others it achieves
${\approx}97\%$. This gives some hope for attention. However, in all
runs, performance on examples with informative open-class lexical items is
no higher than $60\%$, and the systematic errors remain.
It is surprising that the Attention LSTM show this limitation,
as it was designed to overcome the problem of
representational bottlenecks by allowing interaction between the
lexical items in the premise and hypothesis. However,
\newcite{Rock:15} anticipate this failure: ``Word-by-word attention
seems to work well when words in the premise and hypothesis are
connected via deeper semantics or common-sense knowledge'', however
``attention can fail, for example when two sentences and their words
are entirely unrelated''. Our data sets pinpoint this weakness.

The CBoW model also has a bottleneck, but actually performs better
on examples with informative  open-class lexical items than those
without. However, this model performs very poorly on the overall
task of learning complex interactions between function words, so
its performance does not contradict our observation that learning
such interactions results in systematic information loss.

The CompTreeNN avoids this bottleneck by design, since it mixes the
premise and hypothesis via a strict word-by-word alignment. This makes
its learning task much simpler: it need only determine these local
relations and propagate them; we know from \citet{Bowman2:15} that
the relational algebra that underlies this propagation can easily be
learned by standard neural architectures.

\begin{figure}
  \centering
  \input{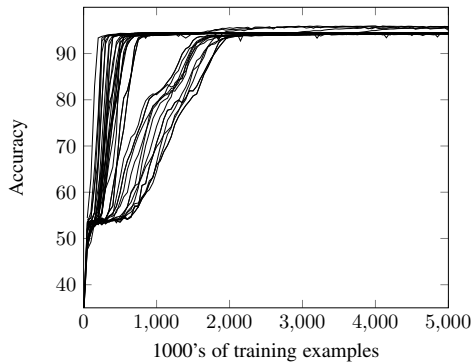}
  \caption{LSTM performance with different hyperparameters. The lower group makes
    decisions based on only quantifiers, negation, and empty words.
    Upper group partly overcomes that suboptimal solution.}
  \label{fig:encoder-model-performance}
\end{figure}

\subsection{The Problem is Architecture}

These systematic errors are not an issue of
low dimensionality; the trends by epoch and final results
are virtually identical with 200-dimensional rather than 100-dimensions
representations.

One might worry that these results represent a failure to optimize
these models properly. We conducted fairly large hyperparameter
searches (section~\ref{sec:models}), but perhaps more optimal settings
lie outside of the space we searched. Figure~\ref{fig:encoder-model-performance}
strongly suggests that this is not the
case. Here, we show the performance of the LSTM with a large
sample of hyperpameter settings. There are two major groups
of mostly indistinguishable runs, those that are completely stuck at
the suboptimal solution and those that partially overcome it.
There is no indication that expanding the hyperparameter search would
change the outcomes for this model and the patterns are the same for
the others we consider. We are left with the conclusion that these
models simply cannot learn to perform this task.

\section{Conclusion}\label{sec:conclusion}

We defined a procedure for generating semantically challenging NLI
data sets and showed that a range of neural models
invariably learn suboptimal solutions that we can characterize based
on the examples themselves. The CompTreeNN overcomes
these limitations, which helps us diagnose the problem: the information
bottleneck formed by learning separate premise and hypothesis
representations. The CompTreeNN is a highly task-specific model, so
its stand-out performance might be seen as more of a challenge
than a victory for deep learning approaches to semantics.

\section*{Acknowledgements}

This material is based in part upon work supported by the
Stanford Data Science Initiative and by the NSF under
Grant No.~BCS-1456077.

\bibliography{naaclhlt2019}
\bibliographystyle{acl_natbib}

\end{document}